\documentclass[10pt,twocolumn]{article} 
\usepackage{simpleConference}
\usepackage{amsmath,graphicx}
\usepackage{xcolor}
\usepackage{amssymb}
\usepackage{algorithm}
\usepackage{algpseudocode}
\usepackage{stackengine}
\usepackage{tabularx,booktabs}
\usepackage{graphicx}
\usepackage{multirow}
\usepackage{hyperref}

\usepackage{xcolor}

\usepackage[nomargin,inline,draft]{fixme}
\fxsetup{theme=color,mode=multiuser}
\FXRegisterAuthor{pf}{apf}{\color{red}pascal}

\DeclareMathOperator*{\argmin}{argmin}

\makeatletter
\usepackage{forloop}
\newcounter{ct}

\usepackage{balance}
\begin{document}

\title{A Relaxed Optimization Approach for Adversarial Attacks against Neural Machine Translation Models }

\author{
Sahar Sadrizadeh$^1$, Clément Barbier$^1$, Ljiljana Dolamic$^2$, Pascal Frossard$^1$ \\\scalebox{0.9}{$^1${\textit{EPFL}}, Lausanne, Switzerland 
\enspace $^2$\textit{Armasuisse S+T}, Thun, Switzerland }
}

\maketitle
\thispagestyle{empty}

\begin{abstract}
In this paper, we propose an optimization-based adversarial attack against Neural Machine Translation (NMT) models. First, we propose an optimization problem to generate adversarial examples that are semantically similar to the original sentences but destroy the translation generated by the target NMT model. This optimization problem is discrete, 
and we propose a continuous relaxation to solve it. With this relaxation, we find a probability distribution for each token in the adversarial example, and then  we can generate multiple adversarial examples by sampling from these distributions. 
Experimental results  show that our attack significantly degrades the translation quality of multiple NMT models while  maintaining the semantic similarity between the original and adversarial sentences. Furthermore, our attack outperforms the baselines in terms of  success rate, similarity preservation, effect on translation quality, and token error rate. Finally, we propose a black-box extension of our attack by sampling from an optimized probability distribution for a reference model whose gradients are accessible\footnote{The source code of our attack can be found at \href{https://github.com/sssadrizadeh/NMT-untargeted-attack}{https://github .com/sssadrizadeh/NMT-untargeted-attack.}}.
\end{abstract}


\section{Introduction}

Neural Machine Translation (NMT) models have been widely used in recent years due to their impressive quality of translation \cite{bahdanau2015neural}. Nevertheless, NMT models are susceptible to carefully crafted perturbations of their input, called adversarial attacks \cite{belinkov2018synthetic,ebrahimi2018adversarial}. Even when the adversarial example is semantically similar to the original sentence, the translation quality of these models may degrade drastically. As adversarial examples  reveal the vulnerabilities of the NMT models, their study is a prerequisite step for making systems robust and reliable. 

Adversarial attacks have been widely studied in computer vision systems \cite{moosavi2016deepfool,ortiz2021optimism}. However, due to the discrete nature of the textual data, the adversarial attack algorithms against image data cannot be directly applied to sentences. For such discrete data (thus not differentiable), it is difficult to employ optimization-based methods, common in computer vision, to generate adversarial examples against NMT systems. 
As studied in \cite{michel2019evaluation}, we define an untargeted adversarial attack against NMT models such that the adversarial example is meaning-preserving in the source language but meaning-destroying in the target language. In other words, the generated adversarial examples should be semantically similar to the original sentences while they decrease translation quality. 

The vulnerability of NMT systems to adversarial attacks has been studied in the literature. First,  Belinkov and Bisk \cite{belinkov2018synthetic} show that character-level NMT models are highly sensitive to substitutions or permutations of letters in input sentences. Furthermore, Ebrahimi et al. \cite{ebrahimi2018adversarial} propose a white-box adversarial attack based on character-level modifications (e.g., flip and insert) by using directional derivatives of the loss of the NMT models to have a first-order estimate of the change in the loss function when a character is changed. Although these character-level attacks show the vulnerability of NMT models, 
they can be easily detected. To generate adversarial examples that are semantically similar to the original sentence, most of the existing adversarial attacks against Natural Language Processing (NLP) systems, and in particular translation systems, consider word replacement \cite{michel2019evaluation,cheng2019robust,zhang2021crafting}. These methods select some of the words in the sentence (randomly or by ranking), then they find suitable substitutions (using a Language Model (LM) or the embedding representation), and finally replace some of the words to fool the target  model. These word-replacement methods may have sub-optimal performance due to their heuristic strategies. Cheng
et al. \cite{cheng2020seq2sick} propose a targeted adversarial attack based on optimization in the embedding space of the NMT model, and they use gradient projection to solve it. Their proposed optimization problem consists of a loss term to fool the target NMT model and a group lasso term to constrain the number of perturbed tokens. Since they use the embedding space of the target NMT model to impose semantic similarity, 
their  adversarial examples may not necessarily be semantically similar to the original sentences. 

As opposed to previous heuristic word-replacement methods, in this paper, we propose an optimization-based method to generate adversarial examples, which are semantically similar to the original sentences, and highly degrade the translation quality of the NMT models. First, we introduce a discrete optimization problem to craft adversarial examples against NMT models. To solve this optimization problem by continuous relaxation, we build upon the idea proposed in \cite{guo2021gradient} and optimize a probability distribution for each token in the adversarial example. By using a language model, we incorporate a differentiable constraint in our optimization problem to ensure the semantic similarity between the original sentence and the adversarial example. 
Afterwards, we can sample from the optimized probability distributions and generate multiple adversarial examples so that we can choose the one that  affects the translation quality the most. 
Moreover, 
we easily extend the proposed method to the black-box setting and show that our attack is effective even if we do not have access to the parameters and gradients of the target model, which may be more dangerous in real-life applications. 
We evaluate our proposed attack against different NMT models and translation tasks. Experimental results indicate that for more than 65\% of sentences, our white-box attack can reduce the translation quality in terms of BLEU score by more than half while maintaining the semantic similarity between the adversarial and original sentences in the source language. Our attack outperforms other white-box and black-box attacks against NMT models in the literature in terms of success rate, decrease in translation quality, semantic similarity, and token error rate. 

The rest of this paper is organized as follows. We formulate the problem of generating adversarial examples against NMT models and the continuous relaxation of the proposed optimization problem  in Section \ref{method}. In Section \ref{experiments}, we 
evaluate our attack against different NMT models in the white-box setting and then extend our attack to the black-box setting. Finally, the paper is concluded in Section \ref{conclusion}.

\section{Proposed Adversarial Attack}  \label{method}

In this section, 
first, we introduce a discrete optimization problem in the token space to find an adversarial example. To solve this optimization problem,  we propose a relaxed continuous optimization problem, which finds a probability distribution for each token in the adversarial example. 

\subsection{Discrete Optimization Problem}
NMT models convert a sequence in the source language to a sequence in the target language. Let $f$ be the target NMT model that maps the input sequence $\mathbf{x}\in \mathcal{X}$ to its translated sequence. We assume that the input sentence is  a sequence of $k$ tokens $\mathbf{x} = x_1x_2...x_k$, where each token comes from a fixed vocabulary $\mathcal{V}$.  The goal of the NMT model is to maximize the probability of the ground-truth translation $\mathbf{y} = y_1y_2...y_m$, that is a sequence of $m$ tokens,  given the input sequence $\mathbf{x}$. We search for an adversarial sentence $\mathbf{x'}$, which we assume to also have $k$ tokens, that fools the target NMT model to generate a translation far from the ground-truth translation. On the other hand, in order to make the perturbation imperceptible, the generated adversarial examples should be  semantically similar to the original sentence in the source language. 

NMT models generate a probability vector over the vocabulary set for each token in the translated sentence. In order to fool the NMT model, we can perturb the input sentence to maximize the training loss (i.e., minimize the negative cross-entropy loss) of the target model. By increasing the cross-entropy loss between the output of the target model and the ground-truth translation, we can maximize the probability of a wrong prediction for the $i$-th token given that the previous tokens have been translated correctly:
\begin{equation}
    \mathcal{L}_{Adv} = - \mathcal{L}_f(\mathbf{x'},\mathbf{y}), \label{eq.ladv}
\end{equation}
where $\mathcal{L}_f$ is  training loss of the NMT model when the input is the adversarial example $\mathbf{x'}$ and  reference translation is $\mathbf{y}$.

In order to maintain semantic similarity between the adversarial and  original sentences, we propose to add a similarity term to the loss function. Many of the works in the literature use embedding representations of the tokens to measure  their similarity \cite{jin2020bert,ren2019generating}. However, if the context is not taken into account, one can produce  unrealistic adversarial examples. We use the contextualized embeddings of an LM to represent the tokens in the original and adversarial sentences $\mathbf{x}$ and $\mathbf{x'}$. The LM, $g$, generates a sequence of contextualized embedding vectors for the input and adversarial sentences as $g(\mathbf{x}) = \langle \mathbf{v}_1,...,\mathbf{v}_k \rangle$ and $g(\mathbf{x'}) = \langle \mathbf{v}'_1,...,\mathbf{v}'_k \rangle$, respectively. The vector representation allows measuring the similarity between two tokens by cosine similarity between their respective vector representations. Thus, we adapt the proposed metric in \cite{zhang2019bertscore}, i.e., BERTScore, to the case of same-length sentences. We compute the similarity loss, $\mathcal{L}_{Sim}$, which measures the distance between the original and adversarial sentence as follows:
\begin{equation}
    \mathcal{L}_{Sim} = -  \sum_{i=1}^{k} w_i \frac{\mathbf{v}_i^\intercal\mathbf{v}'_i}{\|\mathbf{v}_i\|_2.\|{\mathbf{v}'_i}\|_2},
    \label{eq.lsim}
\end{equation}
where $w_i$s are inverse document frequency (idf) scores. Since rare tokens affect the similarity more \cite{vedantam2015cider}, idf scores are used for the importance weighting of the tokens.


Our  optimization problem for generating an adversarial example is a weighted summation of the proposed loss terms: 
\begin{equation} \label{optimization}
    \mathbf{x'} = \argmin_{\substack{\mathbf{x}'_i\in \mathcal{V}}} \;\; [\mathcal{L}_{Adv} + \alpha \mathcal{L}_{Sim}], 
\end{equation}
where $\alpha$ 
is the hyper-parameter that determines the relative importance of the similarity loss term.

\subsection{Continuous Relaxation}

The optimization problem of Eq. \eqref{optimization} is discrete since  the adversarial example should consist of valid  tokens, i.e., the tokens should be in the vocabulary set $\mathcal{V}$. In this section, we propose a continuous relaxation to this optimization problem.

\begin{table*}[t]
	\centering
		\renewcommand{\arraystretch}{0.85}
	\setlength{\tabcolsep}{4pt}
	\caption{Performance of white-box attack against different NMT models
	.}
	\vspace{-5pt}
	\label{tab:white}
	\scalebox{0.9}{
		\begin{tabular}[t]{@{} lcccccccccccccc @{}}
			\toprule[1pt]
		    \multirow{2}{*}{\textbf{Task}}  &
		    \multirow{2}{*}{\textbf{Method}}  & \multicolumn{6}{c}{\textbf{Marian NMT}} &&   \multicolumn{6}{c}{\textbf{mBART50}} \\
			\cline{3-8}
			\cline{10-15}
			\rule{0pt}{2.5ex}    
			& & \scalebox{0.95}{ASR$\uparrow$} & \scalebox{0.95}{RDBLEU$\uparrow$} & \scalebox{0.95}{RDchrF$\uparrow$} & \scalebox{0.95}{Sim.$\uparrow$} & \scalebox{0.95}{Perp.$\downarrow$} & \scalebox{0.95}{TER$\downarrow$} && \scalebox{0.95}{ASR$\uparrow$} & \scalebox{0.95}{RDBLEU$\uparrow$} & \scalebox{0.95}{RDchrF$\uparrow$} & \scalebox{0.95}{Sim.$\uparrow$} & \scalebox{0.95}{Perp.$\downarrow$} & \scalebox{0.95}{TER$\downarrow$}\\
			\midrule[1pt]
			\multirow{3}{*}{En-Fr} 
			& kNN &   \underline{36.24} & \underline{0.35} & \underline{0.17} & 0.82 & 396.45 & 19.32 && \underline{31.22} & \underline{0.29} & 0.12 & \underline{0.85} &  346.28 & 21.12 \\
			& Seq2Sick & 28.11 & 0.21 & 0.16 & 0.74 & \textbf{185.12} & \underline{14.40} && 27.01 & 0.19 & \underline{0.14} & 0.75 &  \textbf{146.01} & \underline{13.68} \\
			& ours &  \textbf{71.79} & \textbf{0.68} & \textbf{0.26} & \textbf{0.85} & \underline{229.06}  & \textbf{12.24} && \textbf{65.16} & \textbf{0.59} & \textbf{0.22} & \textbf{0.87} & \underline{201.61} & \textbf{8.92} \\ 
			\midrule[1pt]
			\multirow{3}{*}{En-De} 
			& kNN &   \underline{39.25} & \underline{0.41} & 0.17 & \underline{0.81} & 445.61 & 19.73 && \underline{36.67} & \underline{0.39} & 0.12 & \textbf{0.85} &  373.26 & 21.49 \\
			& Seq2Sick &  37.34 & 0.31 & \underline{0.21} & 0.66 & \underline{303.60} & \underline{18.70} && 33.54 & 0.29 & \underline{0.19} & 0.66 &  \underline{276.32} & \underline{18.38}\\
			& ours &  \textbf{72.55} & \textbf{0.78} & \textbf{0.26} & \textbf{0.85} &  \textbf{214.47} & \textbf{12.94} && \textbf{66.46} & \textbf{0.71} & \textbf{0.21} & \textbf{0.85} & \textbf{249.72} & \textbf{9.50} \\

			\bottomrule[1pt]
		\end{tabular}
	}
	\vspace{-5pt}
\end{table*}

We can  represent an adversarial sentence with a sequence of  one-hot vectors $\mathbf{z}_i \in \mathbb{R}^{|\mathcal{V}|}$ for each of its tokens. The element of the one-hot vector $\mathbf{z}_i$ that  corresponds  to the index of the token $x_i$ in the vocabulary $\mathcal{V}$ is one, and the other elements are zero.  Therefore, we can optimize the  objective function in \eqref{optimization} with respect to the matrix $\mathbf{Z}  \in \mathbb{R}^{|\mathcal{V}| \times k}$, whose $i$-th column is the one-hot vector $\mathbf{z}_i$. In order to feed a sentence to a transformer model, each token is then transformed into a continuous embedding vector \cite{radford2019language}. We can arrange the input embedding vectors  of the NMT model $f$ for all the tokens in the vocabulary $\mathcal{V}$ as the columns of matrix $\mathbf{E}_f$. 
We also  define matrix $\mathbf{E}_g$, whose columns are the input embedding vectors of the language model $g$. Thus, we can rewrite the optimization problem with respect to $\mathbf{Z}$ as follows:
\begin{equation*}
\scalebox{0.9}{
$
{\argmin\limits_{\mathbf{Z} \in \mathbb{R}^{|\mathcal{V}| \times k}} [\mathcal{L}_{Adv}(\mathbf{E}_f\mathbf{Z}) + \alpha \mathcal{L}_{Sim}(\mathbf{E}_g\mathbf{Z})]}, 
$%
}
\end{equation*}
\begin{equation}
\scalebox{1}{
$
 \mathrm{s.t. } \; \mathbf{Z}_{i,j} \in \{0,1\}, \;  \mathbf{1}_{|\mathcal{V}|}\mathbf{Z} = 1,
$%
}
\end{equation}
where $\mathbf{1}_{|\mathcal{V}|}=[1...1] \in \mathbb{R}^{|\mathcal{V}|}$ is a vector of all ones. The constraints are that $\mathbf{Z}$ is a Boolean matrix and that each column of $\mathbf{Z}$ is a one-hot vector.

In order to  solve this optimization problem, we  consider the continuous relaxation of the one-hot vectors $\mathbf{z}_i$ by using the Gumbel-Softmax trick \cite{jang2017categorical}. In other words,
we  assume that  each token in the adversarial example ($x'_i$)  is a sample from a categorical distribution over the tokens of the vocabulary $\mathcal{V}$: $x'_i \sim \mathrm{Categorical}(\pi_i)$, where $\pi_i = \mathrm{Softmax}(\mathbf{P}_i)$ is the probability distribution over the tokens of the vocabulary. Then, we  use the Gumbel-Softmax trick to sample differentiably from these distributions since it allows the samples to be sparse. Hence, the problem is finding  the probability distributions $\pi_i$, instead of the one-hot vectors $\mathbf{z}_i$.    

However, since the Gumbel-Softmax samples are not one-hot vectors, we approximate the input embedding vectors by computing their linear combination by multiplying the matrix of all the embedding vectors and the Gumbel-Softmax samples. Hence, we can finally reformulate the proposed optimization problem of Eq. \eqref{optimization}:
\begin{equation} \label{optimization-relax}
  \argmin\limits_{\mathbf{P} \in \mathbb{R}^{|\mathcal{V}| \times k}} \mathbb{E}_{\mathbf{Z}'\sim \mathbf{P}'} 
  \scalebox{1}{
$
[\mathcal{L}_{Adv}(\mathbf{E}_f\mathbf{Z}') + \alpha \mathcal{L}_{Sim}(\mathbf{E}_g\mathbf{Z}')], 
$%
}
\end{equation}
where $\mathbf{P}'$ is the Gumbel-Softmax distribution to draw soft samples from the categorical distributions $\pi_i$, and $\mathbf{Z}'$ is a Gumbel-Softmax sample. The optimization problem \eqref{optimization-relax} is continuous. First, we minimize it stochastically by sampling a batch from $\mathbf{P}'$ in each iteration. After optimizing the probability distributions, we  sample from them to generate the tokens of the adversarial example. By sampling multiple times, we can generate different adversarial samples for one input sentence until we find an adversarial example that affects the translation quality more.

\section{Experimental Results}\label{experiments}

In this Section, we evaluate our proposed attack against different NMT models both in white-box and black-box settings. 

\subsection{Setup}

To test our  attack, we  use  En-Fr and En-De validation sets of WMT14  \cite{bojar2014findings}. We also consider different  French-to-English translation datasets from the OPUS corpus collection \cite{tiedemann2012parallel}.  
These datasets are  available on the HuggingFace platform. 

For the target NMT model, we consider Hugging Face implementation \cite{wolf2020transformers} of English-to-French and English-to-German  Marian NMT \cite{bojar2017findings} models  and multilingual mBART50 \cite{tang2020multilingual} NMT model. 
We also need a language model, with the same tokenizer as the target NMT model, for the contextual embeddings in the similarity loss term of our optimization problem. Hence,  we train a language model (with GPT-2 structure \cite{radford2019language})  on the WikiText-103 dataset \cite{merity2016pointer}.

In order to stochastically solve our optimization problem of Eq. \eqref{optimization-relax}, we use an Adam optimizer \cite{kingma2014adam} for 100 iterations with a learning rate of 0.3 and a batch size of 5. The matrix $\mathbf{P}$ is initialized with  zeros except for the elements that correspond to the tokens of the input sentence, which we initialize with a positive value to make them more probable. 
After an ablation study on the coefficients of the optimization problem, we set $\alpha = 45$ for Marian NMT and $\alpha = 120$ for mBART50. Moreover,  we sample 100 times from the optimized Gumbel-Softmax distribution and consider the sample that degrades the translation quality the most, in terms of BLEU score \cite{post-2018-call},  as the generated adversarial example. We evaluate the performance in  terms of different metrics: 1) the attack success rate, where we define an adversarial attack as successful if the BLEU score  is decreased by more than half (ASR); 2) the relative decrease in translation quality in terms of BLEU score  and chrF (RDBLEU and RDchrF); 3) the semantic similarity between the adversarial and clean samples, computed by the multilingual sentence encoder \cite{yang2020multilingual} (Sim.); 4) perplexity score of the adversarial example by GPT-2 (large) language model as a measure of fluency; 5) token error rate of the adversarial example compared to the original sentence (TER). 

Finally,  we compare our attack with other white-box attacks against NMT models in the literature.\footnote{Code of \cite{cheng2019robust}, an untargeted  attack against NMTs, is not publicly available.} We compare our method with the \textit{kNN} attack of \cite{michel2019evaluation}. This method replaces the important tokens in the sentence, found by the gradients, with their nearest tokens in the embedding space of the NMT model. Moreover, we adapt the \textit{Seq2Sick} attack \cite{cheng2020seq2sick}, which is a targeted attack against NMT models, to the untargeted settings considered in our work. Seq2Sick is based on an optimization problem, which contains a similarity term in the NMT embedding space and a group lasso term to impose sparsity  on the number of perturbed tokens.  

\subsection{White-box Attack Results}
The performances of our attack against different NMT models compared to \cite{michel2019evaluation} and \cite{cheng2020seq2sick} can be found in Table \ref{tab:white}. These results are computed for 1000 randomly chosen sentences of the newstest2013 dataset.  We can  see that, on average, our adversarial attack is able to decrease the translation BLEU score by half while maintaining the semantic similarity higher than 0.85. In all cases, for more than 60\% of the input sentences,  our attack generates at least one successful sample, which degrades the BLEU score by more than half. In all cases, our method outperforms other attacks in terms of  semantic similarity, success rate, and token error rate. Moreover, its effect on the translation quality is larger than the baselines in terms of BLEU score and chrF in all cases. The perplexity score of our generated adversarial examples is competitive with those of Seq2Sick.    

Regarding the runtime, in the case of Marian NMT (En-Fr), the optimization of our adversarial distribution takes 2.91 seconds, and the generation of  100 adversarial samples by sampling from the optimized distribution takes 4.75 seconds on a system equipped with an NVIDIA A100 GPU. The kNN attack takes 1.45 seconds to generate an adversarial example; however, it is much less effective than our attack in terms of success rate.  Finally, Seq2Sick takes 38.85  seconds to generate an adversarial example which is much more than  our attack. 

Table \ref{tab:example} shows an  adversarial  example by different attacks against Marian NMT (En-De). As this table shows, our adversarial sample is a fluent sentence and similar to the input sentence. It also largely degrades the BLEU score. However, the changes made by kNN and Seq2Sick are detectable. 

\begin{table}[!t]
	\centering
		\renewcommand{\arraystretch}{1.3}

	\setlength{\tabcolsep}{0.8pt}
	\caption{An Adversarial example against Marian NMT (En-De) generated by different methods. (BLEU score in brackets)}
	\scalebox{0.7}{
		\begin{tabular}{@{}p{1.7cm} | p{10.7cm} @{} }
		
			\midrule[1pt]
			
%
			
		    {Org. } &   During the morning, the Migration and Integration working group also sought to continue its discussions.\hfill\mbox{}  \\ 
			 
			\cline{2-2}
			
			Ref. Trans.  &  Am Vormittag wollte auch die Arbeitsgruppe Migration und Integration ihre Beratungen fortsetzen. \hfill\mbox{} \\ 
			
			\cline{2-2}
			
			Org. Trans. &   Am Vormittag bemühte sich die Arbeitsgruppe "Migration und Integration" auch darum, ihre Diskussionen fortzusetzen. \hfill\mbox{} \\ 
			 
			\cline{1-2}

			{Adv. (ours)} &  During the morning, the Migration and Integration working \textcolor{red}{Players} also sought to continue its discussions. \hfill\mbox{} \\ 
			
			\cline{2-2}
			
			\multirow{2}{*}{Trans.} & Während des Vormittags versuchten die "Migration and Integration Working Players" auch ihre Diskussionen fortzusetzen. \hfill\mbox{} \\ 
			 
			\cline{1-2}
			
			{Adv. (kNN)} & During the morning, the \textcolor{red}{mask} and \textcolor{red}{Integrationos} group also sought \textcolor{red}{with} continue its \textcolor{red}{conversations}. \hfill\mbox{} \\ 
			
			\cline{2-2}
			
			{Trans.} & Am Morgen suchten auch die Maske und Integrationos Gruppe mit ihren Gesprächen weiter. \hfill\mbox{} \\ 

            \cline{1-2}
			
			{Adv. (Seq2Sick)} &  During the \textcolor{red}{early}, the Migration and Integration working group \textcolor{red}{alsoility} to continue its \textcolor{red}{fossil}. \hfill\mbox{} \\ 
			
			\cline{2-2}
			
			\multirow{2}{*}{Trans.} & Während der frühen, die Migration und Integration Arbeitsgruppe auch die Fähigkeit, seine Fossilien fortzusetzen. \hfill\mbox{} \\ 

			\bottomrule[1pt]
		\end{tabular}
	}
	\vspace{-5pt}
	\label{tab:example}
\end{table}

\begin{figure}[tb]
	\centerline{
		\includegraphics[width=0.80\linewidth]{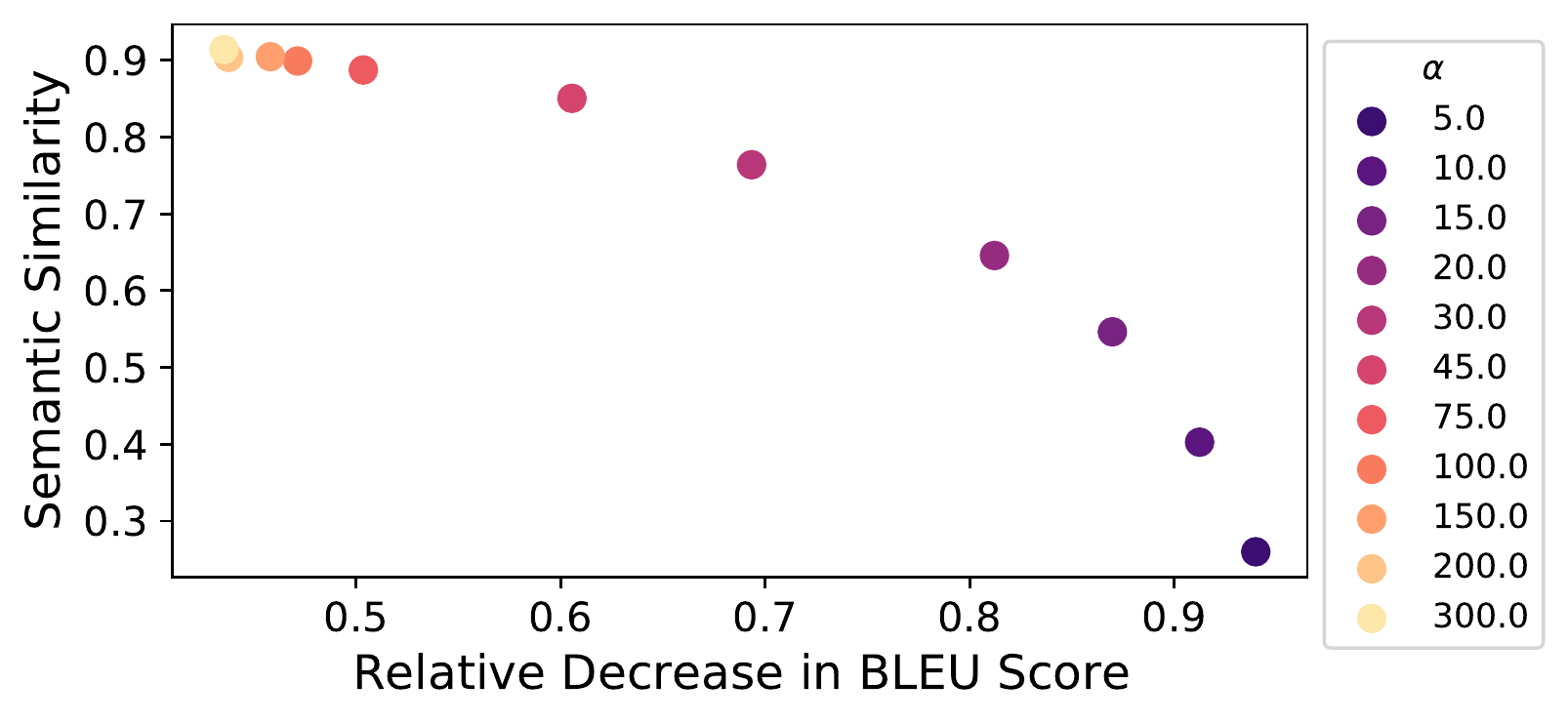}}
	\vspace{-5pt}
	\caption{Effect of the hyper-parameter $\alpha$ on the performance.} 
	\label{fig:sim}
	\vspace{-5pt}
\end{figure}

Furthermore, Fig. \ref{fig:sim} shows the impact of $\alpha$, the coefficient of the similarity loss term in Eq. \eqref{optimization-relax}, on the relative decrease of the BLEU score of the translations of the adversarial and clean sentences, as well as on the semantic similarity between our adversarial example and the original text. These values are computed for Marian NMT (En-Fr) for 200 randomly chosen sentences of the newstest2013 dataset. We can see  that a smaller value of $\alpha$ results in more aggressive adversarial sentences, which  degrade the BLEU score more drastically.

\begin{figure}[tb]
	\centerline{
		\includegraphics[width=0.75\linewidth]{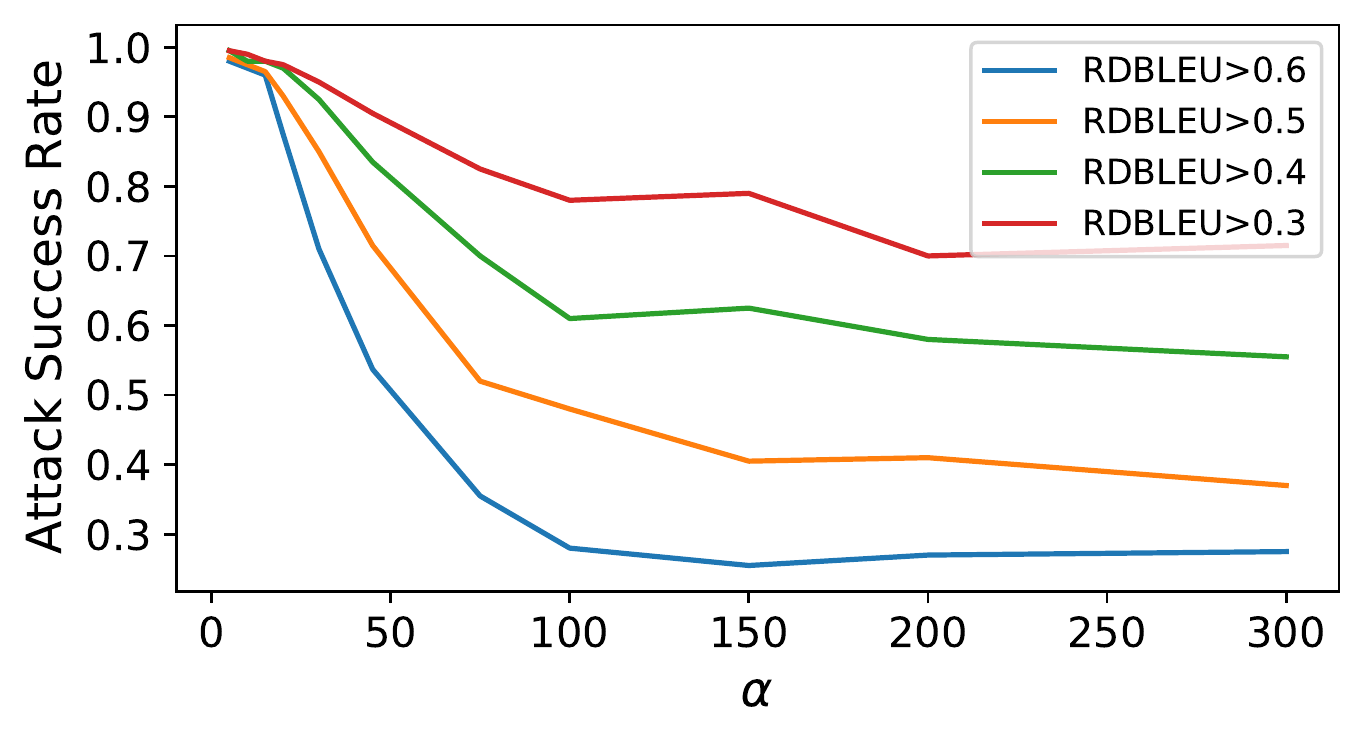}}
	\vspace{-5pt}
	\caption{ASR for different BLEU score ratio thresholds and coefficient  $\alpha$.} 
	\label{fig:success}
	\vspace{-5pt}
\end{figure}

In all the previous results, we considered an attack as successful if the BLEU score is reduced by more than half. Fig. \ref{fig:success} shows the success rate versus different values of $\alpha$  for different thresholds of the  BLEU score ratio criteria. By increasing the similarity coefficient $\alpha$, the attack becomes less aggressive, and the success rate decreases. Moreover, for $\alpha=45$, which was used in our experiments,  the relative decrease in  BLEU score is more  than 0.3 for more than 90\% of the generated adversarial examples.    

To evaluate our  attack  against different types of datasets, we show in Table \ref{tab:white2} the results of our adversarial attack against Marian NMT on 300 randomly chosen sentences from different En-Fr translation datasets from the OPUS corpus collection  \cite{tiedemann2012parallel}. We can see that the performance of our attack is consistent  
among different datasets. The lower performance for OPUS Books in terms of semantic similarity and perplexity can be explained by the fact that this dataset is composed of book passages. Since the nature of this dataset is more distant from Wikitext-103, used to train the language model, the similarity loss term of the optimization problem, which uses the LM contextual embeddings, is less efficient.

\begin{table}[t]
	\centering
		\renewcommand{\arraystretch}{1}
	\setlength{\tabcolsep}{2.7pt}
	\caption{Performance of white-box attack against different datasets.}\vspace{-5pt}
	\label{tab:white2}
	\scalebox{0.83}{
		\begin{tabular}[t]{@{} lcccccc @{}}
			\toprule[1pt]
		    \scalebox{0.95}{\textbf{Datasets}}  & \scalebox{0.95}{ASR$\uparrow$} & \scalebox{0.95}{RDBLEU$\uparrow$} & \scalebox{0.95}{RDchrF$\uparrow$} & \scalebox{0.95}{Sim.$\uparrow$} & \scalebox{0.95}{Perp.$\downarrow$} & \scalebox{0.95}{TER$\downarrow$}\\

			\midrule[1pt]
			 
			 News Commentary & 62.28 & 0.69 & 0.23 & 0.86 & 176.71 & 12.31 \\
			 MultiUN & 61.92 & 0.61 & 0.27 & 0.87 &  159.36 & 12.33\\
			 Europarl & 64.21 & 0.63 & 0.27 & 0.85 & 183.75 & 11.16\\
			 OPUS Books & 70.00 & 0.84 & 0.30 & 0.79 & 377.18 & 14.94 \\
			 
			
			\bottomrule[1pt]
		\end{tabular}
	}
	\vspace{-5pt}
\end{table}

\begin{table}[t]
	\centering
		\renewcommand{\arraystretch}{1}
	\setlength{\tabcolsep}{1.8pt}
	\caption{Performance of  black-box attack against mBART50
	.}
	\vspace{-5pt}
	\label{tab:black}
	\scalebox{0.75}{
		\begin{tabular}[t]{@{} lcccccccc @{}}
			\toprule[1pt]
		    \multirow{1}{*}{\textbf{Task}}  &
		    \multirow{1}{*}{\textbf{Method}}  & 
		    \scalebox{0.95}{ASR$\uparrow$} & \scalebox{0.95}{RDBLEU$\uparrow$} & \scalebox{0.95}{RDchrF$\uparrow$} & \scalebox{0.95}{Sim.$\uparrow$} & \scalebox{0.95}{Perp.$\downarrow$} & \scalebox{0.95}{TER$\downarrow$} &
		    \scalebox{0.95}{\#Queries$\downarrow$}   \\
			\midrule[1pt]
			\multirow{4}{*}{En-Fr} 
			& kNN & 34.44 & 0.33 & 0.15 & 0.82 & 391.49 &  22.66 & - \\
			& Seq2Sick & 25.90 & 0.20 & 0.15 & 0.74 &  \textbf{183.21} & \underline{21.74} &- \\
			& WSLS &   \underline{55.92} &  \underline{0.58} & \textbf{0.27} & \underline{0.84} & 214.30 & 31.12 & 1423 \\
			& ours &   \textbf{66.27} & \textbf{0.59} & \underline{0.23} & \textbf{0.85} & \underline{211.87} & \textbf{14.73} & \textbf{100} \\
			\midrule[1pt]
			\multirow{4}{*}{En-De} 
			& kNN &   37.47 &  0.38 & 0.16 & {0.81} & 435.67 & \underline{22.67} & -\\
			& Seq2Sick & 33.23 & 0.29 & \underline{0.20} & 0.66 & 299.00 & 27.19 & -\\
			& WSLS &   \underline{47.28} &  \underline{0.52} & \underline{0.20} & \textbf{0.86} & \underline{222.88} & 29.09 & 1262\\
			& ours &   \textbf{62.12} & \textbf{0.65} & \textbf{0.20} & \underline{0.85} & \textbf{193.42}  & \textbf{13.22} & \textbf{100} \\
			
			\bottomrule[1pt]
		\end{tabular}
	}
	\vspace{-5pt} 
\end{table}

\subsection{Extension as a Black-box Attack} \label{black}
Since  access to the target model may be limited in real-life applications, it is interesting to analyze the performance of our attack in  black-box settings. To extend our attack to the black-box settings, we propose  to first optimize the distributions of adversarial tokens based on Eq. \eqref{optimization-relax} for a reference NMT model whose gradients are assumed to be accessible. Then for the sampling step of the algorithm, we can sample from these optimized distributions and test them against the target model.  We optimize our adversarial distribution on Marian NMT (since  it is lighter than mBART50) and test the robustness of  mBART50 to the samples taken from those distributions. We compare our attack with WSLS \cite{zhang2021crafting}, a black-box attack against NMT models based on word-replacement. Also, we evaluate kNN and Seq2Sick in the black-box scenario by transferring the adversarial examples, i.e., we attack the target model with the generated adversarial examples against a reference model.  The results are shown in Table \ref{tab:black}. 
We notice  that the performance of our attack is almost unchanged for the black-box setting compared to the white-box setting. Moreover, our attack outperforms all the baselines in  terms of success rate, token error rate, and RDBLEU. Semantic similarity and RDchrF   of our attack and WSLS are also comparable

Regarding the runtime in the black-box settings for English-to-French translation, it takes  2.91 seconds to optimize the probability distributions for the reference model (Marian NMT), and it takes 16.56 seconds to sample 100 adversarial examples from the optimized distributions and test them against the target model (mBART50). On the other hand, WSLS takes 1904.98 seconds to generate an adversarial example, which is far longer than our attack. Finally, the number of queries to the target NMT model required by WSLS is very large, which may not be feasible in practical scenarios.

In summary, our adversarial attack outperforms the baselines in in both white-box and black-box settings. We can generate adversarial examples that preserve the meaning of the original sentences more with a smaller number of perturbed tokens while they highly affect the translation quality.


\section{Conclusion}\label{conclusion}
In this paper, we  proposed an optimization-based adversarial attack against NMT models. We introduced a novel  optimization problem  in the discrete input token space of the target model. Then, we proposed a continuous relaxation by considering a probability distribution for each token in the adversarial example and by using the Gumbel-Softmax trick to differentiably sample from these distributions. We also showed that by sampling from the optimized probability distribution, we can successfully attack target NMT models in the black-box setting. The experimental results demonstrate that the proposed attack is highly effective  against different NMT models and translation tasks. Our attack outperforms other methods in both white-box and black-box scenarios in terms of success rate, semantic similarity in the source language, and token error rate. Our adversarial attack let us better understand the vulnerability of NMT systems and shows the necessity of more robust systems for real-time scenarios. 

\small
\balance
\bibliographystyle{abbrv}
\bibliography{main}

\end{document}